\documentclass[runningheads,a4paper]{llncs}

\usepackage{amssymb}
\usepackage{graphicx}

\usepackage{amsthm}
\usepackage{subfigure}
\usepackage{algorithm}
\usepackage{algorithmic}
\usepackage[bookmarks=true]{hyperref}
\usepackage{multirow}
\usepackage{wrapfig}
\usepackage{enumerate}

\usepackage{url}
\newtheorem{constraint}[theorem]{Constraint}

\begin{document}

\mainmatter  % start of an individual contribution

% first the title is needed
%\title{Antigrasp: a method of detecting antipodal grasps by predicting the presence of necessary geometric conditions}
%\title{Antigrasp: a method of detecting antipodal grasps}
%\title{agile\_grasp: A Method of Detecting Antipodal Grasp Geometries}
%\title{AGILE: A Method of Detecting Antipodal Grasp Geometries}
%\title{A geometry based method of detecting grasps in 3D point clouds}
\title{Using Geometry to Detect Grasps in 3D Point Clouds}

% a short form should be given in case it is too long for the running head
% \titlerunning{Lecture Notes in Computer Science: Authors' Instructions}

% the name(s) of the author(s) follow(s) next

\author{Andreas ten Pas and Robert Platt}

%
% NB: Chinese authors should write their first names(s) in front of
% their surnames. This ensures that the names appear correctly in
% the running heads and the author index.
%
% \author{Alfred Hofmann%
% \thanks{Please note that the LNCS Editorial assumes that all authors have used
% the western naming convention, with given names preceding surnames. This determines
% the structure of the names in the running heads and the author index.}%
% \and Ursula Barth\and Ingrid Haas\and Frank Holzwarth\and\\
% Anna Kramer\and Leonie Kunz\and Christine Rei\ss\and\\
% Nicole Sator\and Erika Siebert-Cole\and Peter Stra\ss er}
% %
% \authorrunning{Lecture Notes in Computer Science: Authors' Instructions}
% (feature abused for this document to repeat the title also on left hand pages)

% the affiliations are given next; don't give your e-mail address
% unless you accept that it will be published
\institute{College of Computer and Information Science, Northeastern University\\
Boston, Massachusetts, USA}

%
% NB: a more complex sample for affiliations and the mapping to the
% corresponding authors can be found in the file "llncs.dem"
% (search for the string "\mainmatter" where a contribution starts).
% "llncs.dem" accompanies the document class "llncs.cls".
%

% \toctitle{Lecture Notes in Computer Science}
% \tocauthor{Authors' Instructions}
\maketitle

\begin{abstract}
This paper proposes a new approach to detecting grasp points on novel
objects presented in clutter. The input to our algorithm is a point
cloud and the geometric parameters of the robot hand. The output is a
set of hand configurations that are expected to be good grasps. Our
key idea is to use knowledge of the geometry of a good grasp to
improve detection. First, we use a geometrically necessary condition
to sample a large set of high quality grasp hypotheses. We were
surprised to find that using simple geometric conditions for detection
can result in a relatively high grasp success rate. Second, we use the
notion of an antipodal grasp (a standard characterization of a good
two fingered grasp) to help us classify these grasp hypotheses. In
particular, we generate a large automatically labeled training set
that gives us high classification accuracy. Overall, our method
achieves an average grasp success rate of 88\% when grasping novels
objects presented in isolation and an average success rate of 73\%
when grasping novel objects presented in dense clutter. This system is
available as a ROS package at \url{http://wiki.ros.org/agile_grasp}.
%\keywords{perception, grasping}
\end{abstract}

%Second, we use geometry to improve our ability to classify hypotheses
%as either good grasps or bad. In particular, we use the notion of an
%antipodal grasp (a standard characterization of a good two fingered
%grasp) to generate large automatically labeled training sets that give
%us high classification accuracy.

\section{Introduction}

Traditionally, robot grasping is understood in terms of two related
subproblems: perception and planning. The goal of the perceptual
component is to estimate the position and orientation (pose) of an
object to be grasped. Then, grasp and motion planners are used to
calculate where to move the robot arm and hand in order to perform
grasp. While this approach can work in ideal scenarios, it has proven
to be surprisingly difficult to localize the pose of novel objects in
clutter accurately~\cite{glover_iros2013}. More recently, researchers
have proposed various grasp point detection methods that localize
grasps independently of object identity. One class of approaches use a
sliding window to detect regions of an RGBD image or a height map
where a grasp is likely to
succeed~\cite{saxena_ijrr2008,jiang_icra2011,fischinger_iros2012,fischinger2013learning,lenz_rss2013,klingbeil_icra2011}. Other
approaches extrapolate local ``grasp prototypes'' based on
human-provided grasp
demonstrations~\cite{detry2013a,herzog_icra2012,kroemer2012kernel}.

A missing element in the above works is that they do not leverage the
geometry of grasping to improve detection. Grasp geometry has been
studied extensively in the literature (for
example~\cite{murray1994mathematical,sudsang}). Moreover, point clouds
created using depth sensors would seem to be well suited for geometric
reasoning. In this paper, we propose an algorithm that detects grasps in a point cloud by predicting
the presence of necessary and sufficient geometric conditions for
grasping. The algorithm has two steps. First, we sample a large set of
grasp hypotheses. Then, we classify those hypotheses as grasps or not
using machine learning. Geometric information is used in both
steps. First, we use geometry to reduce the size of the sample
space. A trivial necessary condition for a grasp to exist is that the
\begin{wrapfigure}{r}{0.5\textwidth}
\begin{center}
%  \subfigure{\includegraphics[height=1.1in]{figs/stereo_setup.png}}
  \subfigure{\includegraphics[height=1.1in]{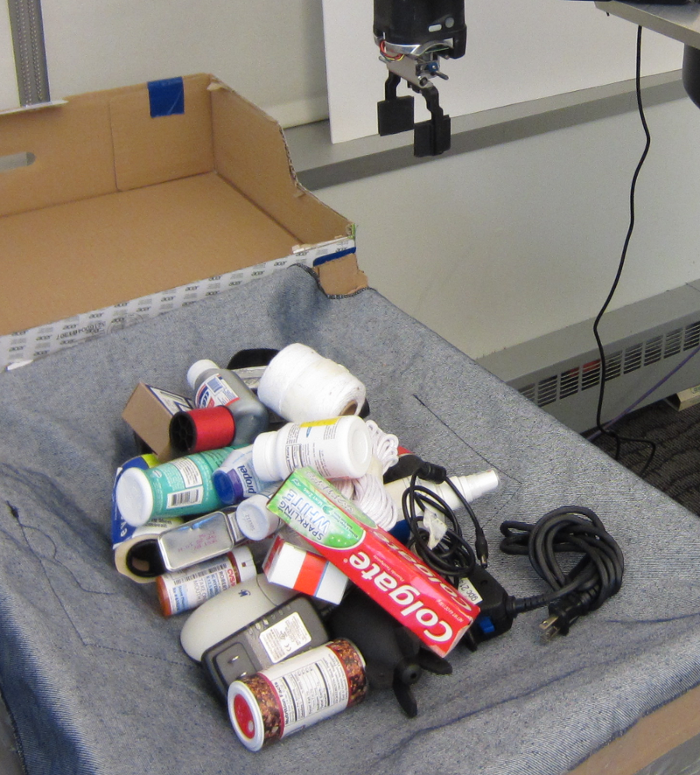}}
  \subfigure{\includegraphics[height=1.1in]{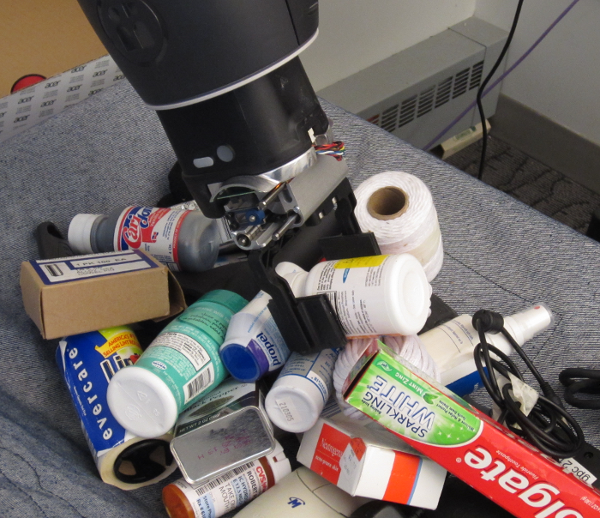}}
\end{center}
  \caption{Our algorithm is able to localize and grasp novel objects
    in dense clutter.}
  \label{fig:overall}
\end{wrapfigure}
hand must be collision-free and part of the object surface must be
contained between the two fingers. We propose a sampling method that
only produces hypotheses that satisfy this condition. This simple step
should boost detection accuracy relative to approaches that consider
every possible hand placement a valid hypothesis. The second way that
our algorithm uses geometric information is to automatically label
the training set. A necessary and sufficient condition for a
two-finger grasp is an antipodal contact configuration (see
Definition~\ref{defn:nguyen}). Unfortunately, we cannot reliably
detect an antipodal configuration in most real-world point clouds
because of occlusions. However, it is nevertheless possible {\em
  sometimes} to verify a grasp using this condition. We use the
antipodal condition to label a subset of grasp hypotheses in arbitrary
point clouds containing ordinary graspable objects. We generate large
amounts of training data this way because it is relatively easy to
take lots of range images of ordinary objects. This is a huge
advantage relative to approaches that depend on human annotations
because large amounts of training data can significantly improve
classification performance.

Our experiments indicate that the approach described above performs
well in practice. We find that without using any machine learning and
just using our collision-free sampling algorithm as a grasp detection
method, we achieve a 73\% grasp success rate for novel objects. This
is remarkable because this is a trivially simple detection
criterion. When a classification step is added to the process, our
grasp success rate jumps to 88\%. This success rate is competitive
with the best results that have been reported. However, what is
particularly interesting is the fact that our algorithm achieves an
average 73\% grasp success rate in dense clutter such as that shown in
Figure~\ref{fig:overall}. This is exciting because dense clutter is a
worst-case scenario for grasping. Clutter creates lots of occlusions
that make perception more difficult and obstacles that make reaching
and grasping harder. 

%The success rate cited above is one of the few experimentally rigorous
%measurements of grasping in clutter that currently exists in the
%literature.

\subsection{Related Work}
\label{sect:related}

The idea of searching an image for grasp targets independently of
object identity was probably explored first in Saxena's early work
that used a sliding window classifier to localize good grasps based on
a broad collection of local visual
features~\cite{saxena_ijrr2008}. Later work extended this concept to
range data~\cite{jiang_icra2011} and explored a deep learning
approach~\cite{lenz_rss2013}. In~\cite{lenz_rss2013}, they obtain an
84\% success rate on Baxter and a 92\% success rate on the PR2 for
objects presented in isolation (averaged over 100 trials). Fischinger
and Vincze developed a similar method that uses heightmaps instead of
range images and develops a different Haar-like feature
representation~\cite{fischinger_iros2012,fischinger2013learning}. In~\cite{fischinger2013learning},
they report a 92\% single-object grasp success rate averaged over 50
grasp trials using the PR2. This work is particularly interesting
because they demonstrate clutter results where the robot grasps and
removes up to 10 piled objects from a box. They report that over six
clear-the-box runs, their algorithm removes an average of 87\% of the
objects from the box. Other approaches search a range image or point
cloud for hand-coded geometries that are expected to be associated
with a good grasp. For example Klingbeil {\em et. al} search a range
image for a gripper-shaped pattern~\cite{klingbeil_icra2011}. In our
prior work, we developed an approach to localizing handles by
searching a point cloud for a cylindrical
shell~\cite{tenpas_iser2014}. Other approaches follow a template-based
approach where grasps that are demonstrated on a set of training
objects are generalized to new objects. For example, Herzog {\em
  et. al} learn to select a grasp template from a library based on
features of the novel object~\cite{herzog_icra2012}. Detry {\em
  et. al} grasp novel objects by modeling the geometry of local object
shapes and fitting these shapes to new
objects~\cite{detry2013a}. Kroemer {\em et. al} propose an object
affordance learning strategy where the system learns to match shape
templates against various actions afforded by those
templates~\cite{kroemer2012kernel}. Another class of approaches worth
mentioning are based on interacting with a stack of objects. For
example, Katz {\em et. al} developed a method of grasping novel
objects based on interactively pushing the objects in order to improve
object segmentation~\cite{katz_icra2013}. Chang {\em et al.} developed
a method of segmenting objects by physically manipulating
them~\cite{chang2012interactive}. The approach presented in this paper
is distinguished from the above primarily because of the way we use
geometric information. Our use of geometry to generate grasp
hypotheses is novel. Moreover, our ability to generate large amounts
of labeled training data could be very important for improving
detection accuracy in the future. However, what is perhaps most
important is that we demonstrate ``reasonable'' (73\%) grasp success
rates in dense clutter -- arguably a worst-case scenario for grasping.

% The approach presented in this paper can be compared directly with the
% above work. In this paper, we report a 96\% success rate averaged over
% 115 grasps of 31 novel objects in a single-object setting. We also
% report clutter experiments where we remove 90\% of the objects from a
% pile of 26 objects. These results are impressive for two
% reasons. First, a 96\% success rate is higher than anything else
% reported above. Second, none of the work reported above considered a
% clutter scenario with as many as 26 objects.

\section{Approach}
\label{sect:approach}

We frame the problem of localizing grasp targets in terms of locating
{\em antipodal hands}, an idea that we introduce based on the concept
of an antipodal grasp. In an antipodal grasp, the robot hand is able
to apply opposite and co-linear forces at two
points:

\begin{definition}[Nguyen~\cite{nguyen_icra86}]
\label{defn:nguyen}
A pair of point contacts with friction is {\bf antipodal} if and only
if the line connecting the contact points lies inside both friction
cones~\footnote{A friction cone describes the space of normal and
  frictional forces that a point contact with friction can apply to
  the contacted surface~\cite{murray1994mathematical}.}.
\end{definition}

\noindent
If an antipodal grasp exists, then the robot can hold the object by
applying sufficiently large forces along the line connecting the two
contact points. In this paper, we restrict consideration to parallel
jaw grippers -- hands with parallel fingers and a single closing
degree of freedom. Since a parallel jaw gripper can only apply forces
along the (single) direction of gripper motion, we will additionally
require the two contact points to lie along a line parallel to the
direction of finger motion. Rather than localizing antipodal contact
configurations directly, we will localize hand configurations where we
expect an antipodal grasp to be achieved in the future when the hand
closes. Let $\mathcal{W} \subseteq \mathbb{R}^3$ denote the robot
workspace and let $\mathcal{O} \subseteq \mathcal{W}$ denote space
occupied by objects or obstacles. Let $H \subseteq SE(3)$ denote the
configuration space of the hand when the fingers are fully open. We
will refer to a configuration $h \in H$ as simply a ``hand''. Let
$B(h) \subseteq W$ denote the volume occupied by the hand in
configuration $h \in H$, when the fingers are fully open.

%Let $\hat{f}(h)$ denote the direction of closing of one finger. (In a
%parallel jaw gripper, the other finger closes in the opposite
%direction).

\begin{definition}\label{def:pre_antipodal_hand}
An {\bf antipodal hand} is a pose of the hand, $h \in H$, such
that the hand is not in collision with any objects or obstacles, $B(h)
\cap \mathcal{O} = \emptyset$, and at least one pair of antipodal
contacts will be formed when the fingers close such that the line
connecting the two contacts is parallel to the direction of finger
motion.
\end{definition}

%\noindent
%In practice, we search for hands that are nearly antipodal. This
%approximation is designed to allow for small shifting in object pose
%during grasp formation.

%\begin{definition}\label{def:antipodal_hand}
%A {\bf $\theta$-near antipodal hand} is a hand, $h_n \in H$, for which
%there exists an antipodal hand, $h \in H$, that can be obtained by
%rotating $h_n$ by no more than $\theta$ about a reference point, $r
%\in R(h_n)$.
%\end{definition}

Algorithm~\ref{alg:overall} illustrates at a high level our algorithm
for detecting antipodal hands. It takes a point cloud, $\mathcal{C}
\subseteq \mathbb{R}^3$, and a geometric model of the robot hand as
input and produces as output a set of hands, $\mathcal{H} \subseteq
H$, that are predicted to be antipodal. There are two main
steps. First, we sample a set of hand hypotheses. Then, we classify
each hypothesis as an antipodal hand or not. These steps are described
in detail in the following sections.

\begin{algorithm}
\caption{Detect\_Antipodal\_Hands}
\vspace{0.05in}
{\bf Input:} a point cloud, $\mathcal{C}$, and hand parameters, $\theta$\\
{\bf Output:} antipodal hands, $\mathcal{H}$\\
\vspace{-0.15in}
\label{alg:overall}
\begin{algorithmic}[1]
\STATE $\mathcal{H}_{hyp} = Sample\_Hands(\mathcal{C})$
\STATE $\mathcal{H} = Classify\_Hands(\mathcal{H}_{hyp})$
\end{algorithmic}
\end{algorithm}

\section{Sampling Hands}
\label{sect:finding_hypotheses}

A key part of our algorithm is the approach to sampling from the space
of hand hypotheses. A naive approach would be to sample directly from
$H \subseteq SE(3)$. Unfortunately, this would be immensely
inefficient because $SE(3)$ is a 6-DOF space and many hands sampled
this way would be far away from any visible parts of the point
cloud. Instead, we define a lower-dimensional sample space constrained
by the geometry of the point cloud.

%Instead, we introduce a constrained sample space that focuses sampling
%on likely antipodal grasps.

%Instead, we introduce two sampling constraints that help to focus
%sampling on likely antipodal grasps.
\pagebreak

%\subsection{Geometry of objects and the hand}
\subsection{Geometry of the Hand and the Object Surface}

\begin{wrapfigure}{r}{0.5\textwidth}
\begin{center}
  \subfigure[]{\includegraphics[height=0.9in]{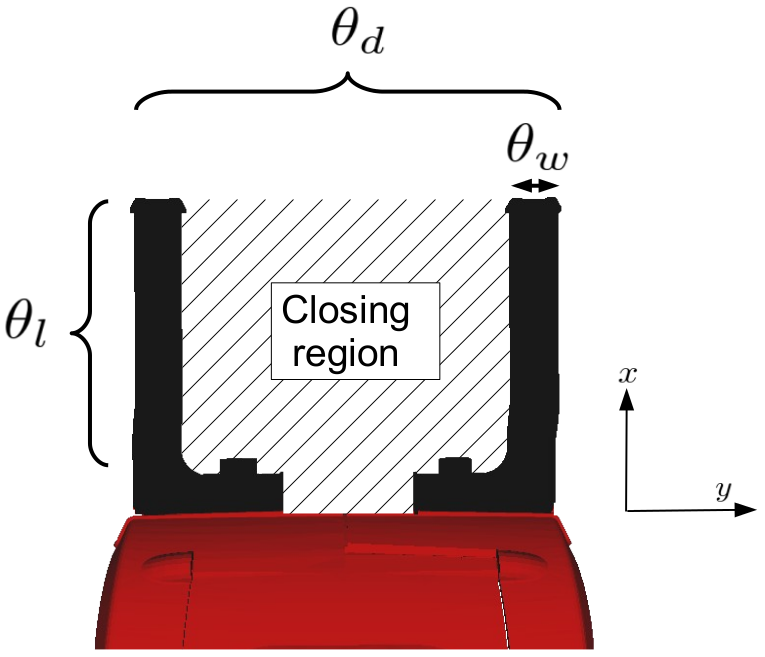}}
  \subfigure[]{\includegraphics[height=0.9in]{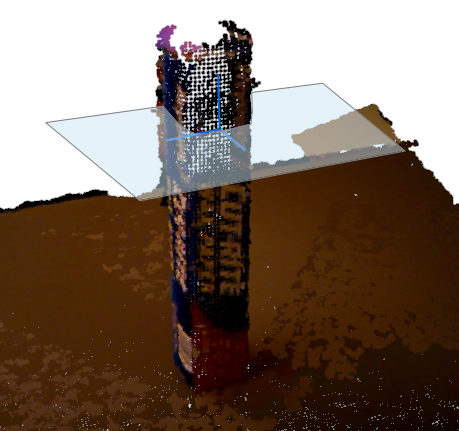}}
\end{center}
  \caption{(a) hand geometry. (b) cutting plane geometry.}
  \label{fig:hand}
\end{wrapfigure}

Before describing the sample space, we quantify certain parameters
related to the grasp geometry. We assume the hand, $h \in H$, is a
parallel jaw gripper comprised of two parallel fingers each modeled as
a rectangular prism that moves parallel to a common plane. Let
$\hat{a}(h)$ denote a unit vector orthogonal to this plane. The hand
is fully specified by the parameter vector $\theta = (\theta_l,
\theta_w, \theta_d, \theta_t)$ where $\theta_l$ and $\theta_w$ denote
the length and width of the fingers; $\theta_d$ denotes the distance
between the fingers when fully open; and $\theta_t$ denotes the
thickness of the fingers (orthogonal to the page in
Figure~\ref{fig:hand} (a)). Define the {\bf closing region}, $R(h)
\subseteq \mathcal{W}$, to be the volumetric region swept out by the
fingers when they close. Let $r(h) \in R(h)$ denote an arbitrary
reference point in the closing region. Define the {\bf closing plane},
$C(h)$, to be the subset of the plane that intersects $r(h)$, is
orthogonal to $\hat{a}(h)$, and is contained within $R(h)$:
\[
C(h) = \{p \in R(h) | (p-r(h))^T\hat{a}(h) = 0\}.
\]

%Define the {\bf closing plane} to be the plane that passes
%through $r(h)$ and is parallel to the plane in which the fingers
%move. Let $\hat{a}(h)$ denote the axis orthogonal to the closing
%plane. 

%two fingers always move through a single plane that we will call the
%{\bf closing plane} (parallel to the page in
%Figure~\ref{fig:hand}). We will denote the normal vector orthogonal to
%this plane as $\hat{a}(h)$.

We also introduce some notation related to the differential geometry
of the surfaces we are grasping. Recall that each point on a
differentiable surface is associated with a surface normal and two
principal curvatures where each principal curvature is associated with
a principal direction. The surface normal and the two principal
directions define an orthogonal basis known as a Darboux
frame~\footnote{Any frame aligned with the surface normal is a Darboux
  frame. Here we restrict consideration to the special case where it
  is also aligned with the principal directions.}. The Darboux frame
at point $p \in \mathcal{C}$ will be denoted: $F(p) = (\hat{n}(p) \;
(\hat{a}(p) \times \hat{n}(p)) \; \hat{a}(p))$, where $\hat{n}(p)$
denotes the unit surface normal and $\hat{a}(p)$ denotes the direction
of minimum principal curvature at point $p$. Define the {\bf cutting
  plane} to be the plane orthogonal to $\hat{a}(p)$ that passes
through $p$ (see Figure~\ref{fig:hand} (b)). Since we are dealing with
point clouds, it is not possible to measure the Darboux frame exactly
at each point. Instead, we estimate the surface normal and principle
directions over a small neighborhood. We fit a quadratic function over
the points contained within a small ball (3 cm radius in our
experiments) using Taubin's
method~\cite{taubin_pami1991,tenpas_iser2014} and use that to
calculate the Darboux frame~\footnote{Taubin's method is an analytic
  solution that performs this fit efficiently by solving a generalized
  Eigenvalue problem on two $10\times10$
  matrices~\cite{taubin_pami1991}. In comparison to using first order
  estimates of surface normal and curvature, the estimates derived
  from this quadratic are more robust to local surface
  discontinuities.}.

\subsection{Hand Sample Set}

We want a set that contains many antipodal hands and from which it is
easy to draw samples. The following conditions define the set
$\mathcal{H}$. First, for every hand, $h \in \mathcal{H}$:

\begin{constraint}
The body of the hand is not in collision with the point cloud: $B(h)
\cap \mathcal{C} = \emptyset$,
\end{constraint}

\noindent
Furthermore, there must exist a point in the cloud, $p \in
\mathcal{C}$, such that:

\begin{constraint}
The hand closing plane contains $p$: $p \in C(h)$.
\end{constraint}

\begin{constraint}
\label{const:parallel}
The closing plane of the hand is parallel to the cutting plane at $p$:
$\hat{a}(p) = \hat{a}(h)$.
\end{constraint}

\noindent
These three constraints define the following set of hands:
\begin{equation}
\mathcal{H} = \cup_{p \in \mathcal{C}} H(p), \; H(p) = \{h \in H | p \in C(h) \wedge \hat{a}(p) = \hat{a}(h) \wedge B(h) \cap \mathcal{C} = \emptyset \}.
\label{eqn:sampleset}
\end{equation}

\noindent
Constraint~\ref{const:parallel} is essentially a heuristic that limits
the hand hypotheses that our algorithm considers. While this
eliminates from consideration many otherwise good grasps, it is a
practical way to focus detection on likely candidates. Moreover, it is
easy to sample from $\mathcal{H}$ by: 1) sampling a point, $p \in
\mathcal{C}$, from the cloud; 2) sampling one or more hands from
$H(p)$. Notice that for each $p \in \mathcal{C}$, $H(p)$ is three-DOF
because we have constrained two DOF of orientation and one DOF of
position. This means that $\mathcal{H}$ is much smaller than $H$ and
it can therefore be covered by many fewer samples.

\begin{algorithm}
\caption{Sample\_Hands}
\vspace{0.05in}
{\bf Input:} point cloud, $\mathcal{C}$, hand parameters, $\theta$\\
{\bf Output:} grasp hypotheses, $\mathcal{H}$\\
\vspace{-0.15in}
\label{alg:gethypo}
\begin{algorithmic}[1]
\STATE $\mathcal{H} = \emptyset$
\STATE Preprocess $\mathcal{C}$ (voxelize; workspace limits; {\em etc.})
\FOR{i = 1 to n} 
\STATE Sample $p \in \mathcal{C}$ uniformly randomly
\STATE Calculate $\theta_d$-ball about $p$: $N(p) = \{q \in \mathcal{C}
: \|p - q\| \leq \theta_d\}$
\STATE Estimate local Darboux frame at $p$: $F(p) = Estimate\_Darboux(N(p))$
\STATE $H = Grid\_Search(F(p),N(p))$
\STATE $\mathcal{H} = \mathcal{H} \cup H$
\ENDFOR
\end{algorithmic}
\end{algorithm}

The sampling process is detailed in
Algorithm~\ref{alg:gethypo}. First, we preprocess the point cloud,
$\mathcal{C}$, in the usual way by voxelizing (we use voxels 3mm on a
side in our experiments) and applying workspace limits (Step
2). Second, we iteratively sample a set of $n$ points ($n$ is between
4000 and 8000 in our experiments) from the cloud (Step 4). For each
point, $p \in \mathcal{C}$, we calculate a neighborhood, $N(p)$, in
the $\theta_d$-ball around $p$ (using a KD-tree, Step 5). The next
step is to estimate the Darboux frame at $p$ by fitting a quadratic
surface using Taubin's method and calculating the surface normal and
principal curvature directions (Step 6). Next, we sample a set of hand
configurations over a coarse two-DOF grid in a neighborhood about
$p$. Let $h_{x,y,\phi}(p) \in H(p)$ denote the hand at position
$(x,y,0)$ with orientation $\phi$ with respect to the Darboux frame,
$F(p)$. Let $\Phi$ denote a discrete set of orientations (8 in our
implementation). Let $X$ denote a discrete set of hand positions (20
in our implementation). For each hand configuration $(\phi,x) \in \Phi
\times X$, we calculate the hand configuration furthest along the $y$
axis that remains collision free: $y^* = \max_{y \in Y}$ such that
$B(h_{x,y,\phi}) \cap N = \emptyset$, where $Y=[-\theta_d,\theta_d]$
(Step 3). Then, we check whether the closing plane for this hand
configuration contains points in the cloud (Step 4). If it does, then
we add the hand to the hypotheses set (Step 5).

%We exhaustively search $H(p)$ over a coarse two dimensional grid.

%and exhaustively search $H(p)$ for a discrete set of hands that
%satisfy it. This is accomplished using exhaustive search (Step 7 of
%Algorithm~\ref{alg:gethypo}), as described in the next subsection.

%Our experiments bear out this intuition: depending on scene
%complexity, we have found this method to generate between 500 (for a
%scene containing a single object lying on a table) and 2000 hand
%hypotheses (for a densely cluttered scene).

%\begin{figure}[t]
%\begin{center}
%  \subfigure[]{\includegraphics[height=1.1in]{figs/handdimensions4.png}}
%\hspace{0.1in}
%  \subfigure[]{\includegraphics[height=1.1in]{figs/cutrite_pic.png}}
%\hspace{0.1in}
%  \subfigure[]{\includegraphics[height=1.1in]{figs/cutrite_plane2.png}}
%\hspace{0.1in}
%  \subfigure[]{\includegraphics[height=1.1in]{figs/pointinplane2.png}}
%\end{center}
%  \caption{(a) Hand geometry; (b) Camera image of a typical
%    object; (c) Illustration of the cutting plane for a point, $s_i
%    \in \mathcal{C}$, on the surface of the object; (d) Neighborhood
%    points projected onto the cutting plane, $\Lambda(s_i)$.}
%  \label{fig:cuttingplane}
%\end{figure}

%\subsection{Exhaustive Search over $H(p)$}

\begin{algorithm}
\caption{Grid\_Search}
\vspace{0.05in}
{\bf Input:} neighborhood point cloud, $N$; Darboux frame, $F$\\
{\bf Output:} neighborhood grasp hypotheses, $H$\\
\vspace{-0.15in}
\label{alg:grasphyp}
\begin{algorithmic}[1]
\STATE $H = \emptyset$
\FORALL{$(\phi,x) \in \Phi \times X$} 
\STATE Push hand until collision: $y^* = \max_{y \in Y}$ such that $B(h_{\phi,x,y}) \cap N = \emptyset$
\IF{closing plane not empty: $C(h_{\phi,x,y^*}) \cap N \neq \emptyset$}
\STATE {$H = H \cup h_{\phi,x,y^*}$}
\ENDIF
\ENDFOR
\end{algorithmic}
\end{algorithm}

%This approach effectively generates lots of antipodal hand
%hypotheses. Depending on scene complexity, we have found this method
%to generate between 500 (for a scene containing a single object lying
%on a table) and 2000 hand hypotheses (for a densely cluttered scene).

\subsection{Grasping Results}

Interestingly, our experiments indicate that this sampling method by
itself can be used to do grasping. In Algorithm~\ref{alg:overall}, the
sampling process is followed by the grasp classification process
described in the next section. However, if we omit classification,
implicitly assuming that all grasp hypotheses are true grasps, we
obtain a surprisingly high grasp success rate of approximately 73\%
(the column labeled {\em NC, 2V} in
Figure~\ref{fig:per_obj_results_all}). The experimental context of
this result is described in Section~\ref{sect:exps}. Essentially, we
cluster the sampled hands and use a heuristic grasp selection strategy
to choose a grasp to execute (see
Section~\ref{sect:grasp_selection}). This result is surprising because
the sampling constraints (Constraints 1--3) encode relatively simple
geometric conditions. It suggests that these sampling constraints are
an important part of our overall grasp success rates.

\section{Classifying Hand Hypotheses}

After generating hand hypotheses, the next step is to classify each of
those hypotheses as antipodal or not. The simplest approach would be
to infer object surface geometry from the point cloud and then check
which hands satisfy
Definition~\ref{def:pre_antipodal_hand}. Unfortunately, since most
real-world point clouds are partial, many hand hypotheses will fail
this check simply because all relevant object surfaces were not
visible to a sensor. Instead, we infer which hypotheses are likely to
be antipodal using machine learning ({\em i.e.}  classification).

\subsection{Labeling Grasp Hypotheses}
\label{sect:identify_antipodal}

\begin{wrapfigure}{r}{0.3\textwidth}
\begin{center}
  \includegraphics[width=1.3in]{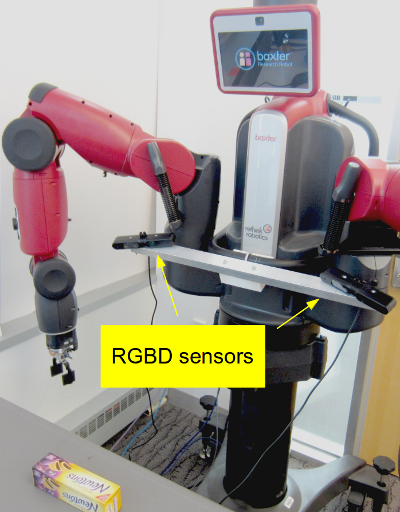}
\end{center}
  \caption{Our robot has stereo RGBD sensors.}
  \label{fig:stereo}
\end{wrapfigure}

%Obtaining labeled training data is a major challenge in machine
%learning. 

Many approaches to grasp point detection require large amounts of
training data where humans have annotated images with good grasp
points~\cite{saxena_ijrr2008,jiang_icra2011,lenz_rss2013,fischinger_iros2012,fischinger2013learning,herzog_icra2012}. Unfortunately,
obtaining these labels is challenging because it can be hard for human
labelers to predict what object surfaces in a scene might be graspable
for a robot. Instead, our method automatically labels a set of
training images by checking a relaxed version of the conditions of
Definition~\ref{def:pre_antipodal_hand}.
%Instead, we propose an approach to automatically labeling
%training images with good grasp points. Because of the way we
%structure the problem, it is possible in some cases to positively
%determine that certain hand hypothese are antipodal or not by 

In order to check whether a hand hypotheses, $h \in H$, is antipodal,
we need to determine whether an antipodal pair of contacts will be
formed when the hand closes. Let $\hat{f}(h)$ denote the direction of
closing of one finger. (In a parallel jaw gripper, the other finger
closes in the opposite direction). When the fingers close, they will
make first contact with an extremal pair of points, $s_1, s_2 \in
R(h)$ such that $\forall s \in R(h), s_1^T \hat{f}(h) \geq s^T
\hat{f}(h) \wedge s_2^T \hat{f}(h) \leq s^T \hat{f}(h)$. An antipodal
hand requires two such extremal points to be antipodal and for the
line connecting the points to be parallel to the direction of finger
closing. In practice, we relax this condition slightly as
follows. First, rather than checking for extremal points, we check for
points that have a surface normal parallel to the direction of
closing. This is essentially a first-order condition for an extremal
point that is more robust to outliers in the cloud. The second way
that we relax Definition~\ref{def:pre_antipodal_hand} is to drop the
requirement that the line connecting the two contacts be parallel to
the direction of finger closing and to substitute a requirement that
at least $k$ points are found with an appropriate surface
normal. Again, the intention here is to make detection more robust: if
there are at least $k$ points near each finger with surface normals
parallel to the direction of closing, then it is likely that the line
connecting at least one pair will be nearly parallel to the direction
of finger closing. In summary, we check whether the following
definition is satisfied:

\begin{definition}
\label{defn:nearantipodal}
A hand, $h \in H$, is {\bf near antipodal} for thresholds $k \in
\mathbb{N}$ and $\theta \in [0,pi/2]$ when there exist $k$ points
$p_1, \dots, p_k \in R(h) \cap \mathcal{C}$ such that
$\hat{n}(p_i)^T\hat{f}(h) \geq \cos \theta$ and $k$ points $q_1,
\dots, q_k \in R(h) \cap \mathcal{C}$ such that
$\hat{n}(q_i)^T\hat{f}(h) \leq -\cos \theta$.
\end{definition}

\noindent
When Definition~\ref{defn:nearantipodal} is satisfied, then we label
the corresponding hand a positive instance. Note that in order to
check for this condition, it is necessary to register at least two
point clouds produced by range sensors that have observed the scene
from different perspectives (Figure~\ref{fig:stereo}). This is because
we need to ``see'' two nearly opposite surfaces on an object. Even
then, many antipodal hands will not be identified as such because only
one side of the object is visible. These ``indeterminate'' hands are
omitted from the training set. In some cases, it is possible to verify
that a particular hand is {\em not} antipodal by checking that there
are fewer than $k$ points in the hand closing region that satisfy
either of the conditions of Definition~\ref{defn:nearantipodal}. These
hands are included in the training set as negative examples. This
assumes that the closing region of every sampled hand hypothesis is at
least partially visible to a sensor. If there are fewer than $k$
satisfying points, then Definition~\ref{defn:nearantipodal} would not
be satisfied even if the opposite side of an object was observed. In
our experiments, we set the thresholds $k=6$ and $\theta=20$ degrees.

%In creating the training set, it is important to ensure the negative
%examples are accurate. The problem is that many hand hypotheses in the
%point clouds used to create the training data will fail satisfy
%Definition~\ref{def:antipodal_hand} as a result of incomplete data. We
%need a way to ensure that every negative instance {\em cannot} satisfy
%Definition~\ref{def:antipodal_hand} regardless of additional
%data. Fortunately, this is easy to check. Any hypothesis for which
%neither of its constituent fingers is antipodal cannot be an antipodal
%hand and is therefore labeled as a negative instance. Any hypotheses
%with one antipodal finger but not two is considered to be an {\bf
%  indeterminate} hand -- it could potentially be found to be antipodal
%given more point cloud data.

\subsection{Feature Representation}
\label{sect:feature}

\begin{wrapfigure}{r}{0.25\textwidth}
%\begin{center}
  \includegraphics[height=1.0in]{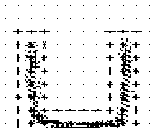}
%\end{center}
  \caption{HOG feature representation of a hand hypothesis for the box shown in Figure~\ref{fig:hand} (b).}
  \label{fig:hog}
\end{wrapfigure}

%Let $F(h) = \left(\hat{x}(h) \hat{y}(h)\right)$
%denote the reference frame of the $x$ and $y$ directions (see
%Figure~\ref{fig:cuttingplane} (a)) in the base reference frame. 

In order to classify hand hypotheses, a feature descriptor is
needed. Specifically, for a given hand $h \in H$, we need to encode
the geometry of the points contained within the hand closing region,
$\mathcal{C} \cap R(h)$. A variety of relevant descriptors have been
explored in the
literature~\cite{knopp2010hough,rusu_icra2009,tombari2010unique}. In
our case, we achieve good performance using a simple descriptor based
on HOG features. For a point cloud, $\mathcal{C}$, a two dimensional
image of the closing region is created by projecting the points
$\mathcal{C} \cap R(h)$ onto the hand closing plane: $I(\mathcal{C},h)
= S_{12} F(h)^T \left( N \cap C(h) \right)$, where $S_{12} = \left(
\begin{array}{ccc}
1 & 0 & 0 \\
0 & 1 & 0
\end{array}
 \right)$ selects the first two rows of $F(h)^T$. We call this the
       {\bf grasp hypothesis image}. We encode it using the HOG
       descriptor, $HOG(I(\mathcal{C},h))$. In our implementation, we
       chose a HOG cell size such that the grasp hypothesis image was
       covered by $10 \times 12$ cells with a standard $2 \times 2$
       block size.

%We chose the HOG descriptor~\cite{dalal2005histograms} because it is
%easy to use, but there are certainly lots of
%alternatives~\cite{viola2001rapid,felzenszwalb2010object}. Figure~\ref{fig:hog}
%illustrates the HOG representation corresponding to the given grasp
%hypothesis image. We have found that our method is not terribly
%sensitive to the parameters of the HOG descriptor.

%The goal with these features is typically to describe a very small
%neighborhood uniquely in order to do registration. Our intent is
%different. We want to encode the geometry of the object surface
%contained in the closing region of the grasp hypothesis. Our
%descriptor is inspired by the method that we used to sample grasp
%hypotheses. Recall that we projected local neighborhood points onto
%the cutting plane and that the coordinate frame of the cutting plane
%was aligned with the surface normal (Step 6 of
%Algorithm~\ref{alg:grasphyp}). Our approach is to convert this cutting
%plane projection into an image and apply standard 2D image
%classification techniques. In particular, we will convert the portion
%of the cutting plane that lies within the closing region of the hand
%into an image called {\bf grasp hypothesis image}. There are a variety
%of potential classification methods that could be used. In this case,
%since we want something that will generalize appropriately across
%similarly shaped object surfaces, we want a method designed for object
%category recognition.

\begin{figure}[t]
\begin{center}
  \subfigure[]{\includegraphics[height=0.9in]{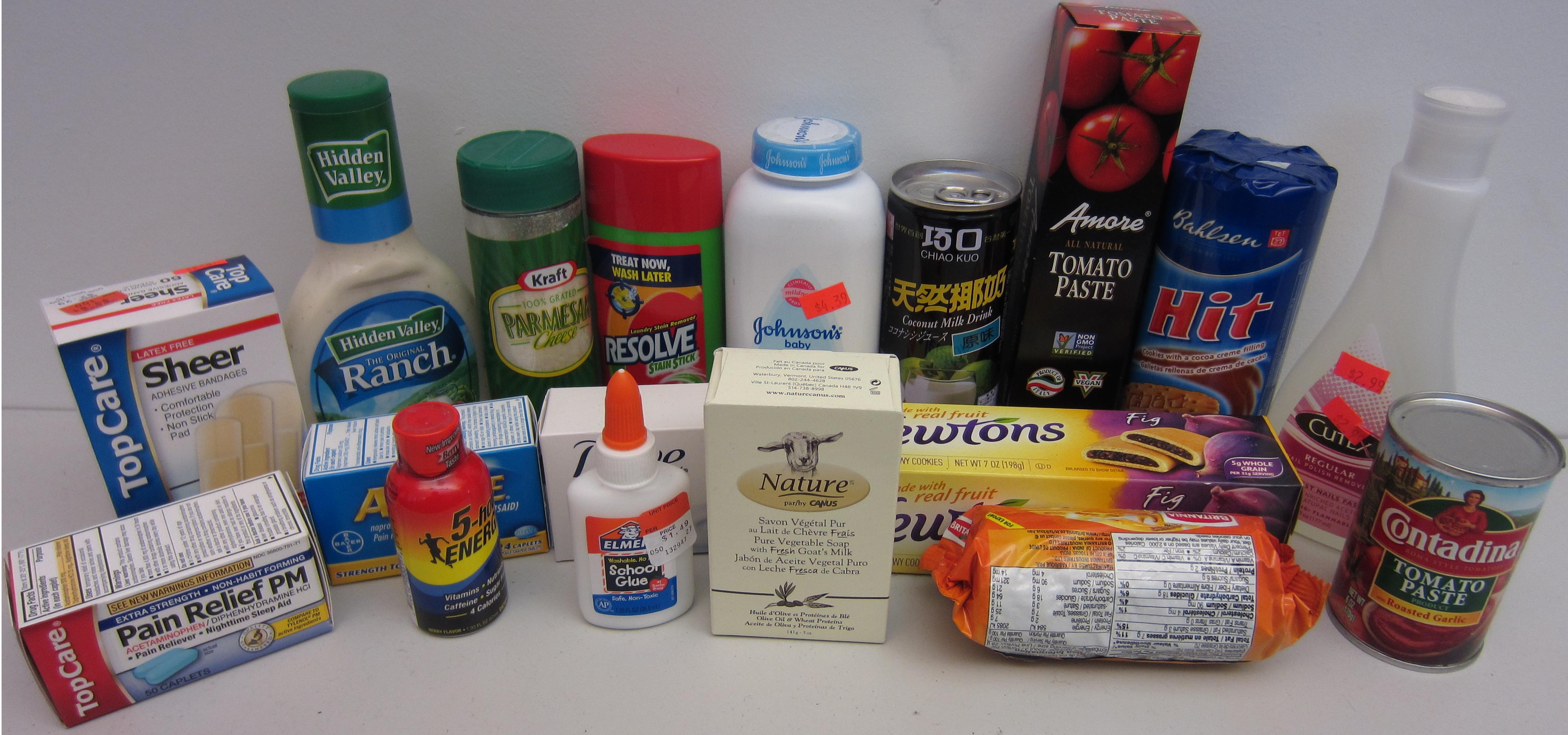}}
\hspace{0.1in}
  \subfigure[]{\includegraphics[height=1.0in]{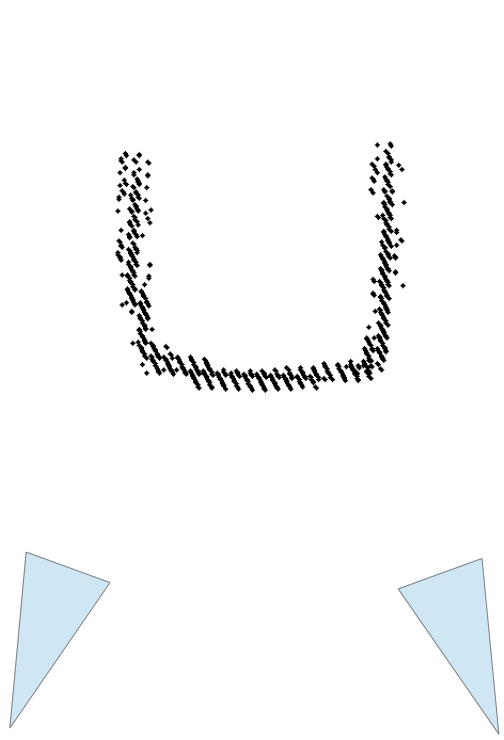}}
\hspace{0.1in}
  \subfigure[]{\includegraphics[height=1.0in]{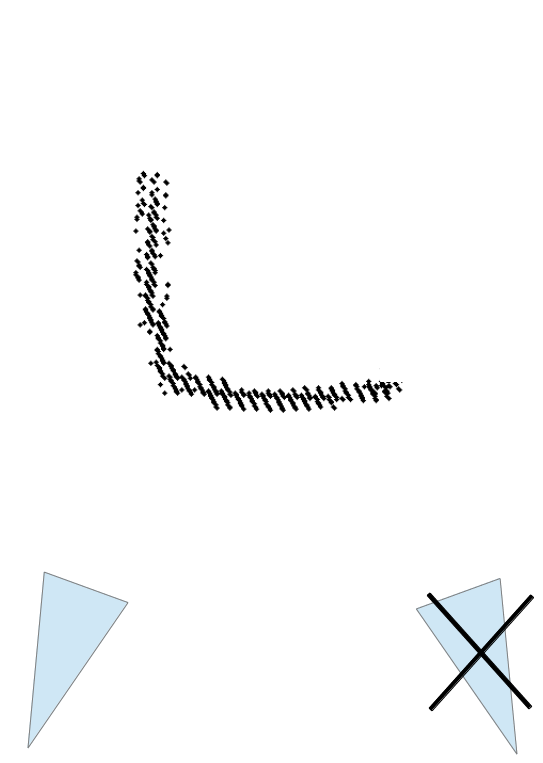}}
\hspace{0.1in}
  \subfigure[]{\includegraphics[height=1.0in]{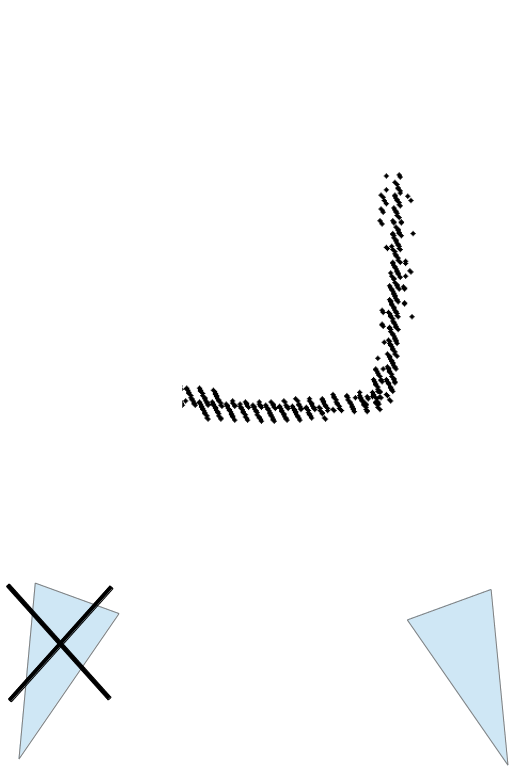}}
\end{center}
  \caption{(a) training set comprised of 18 objects. (b-d)
    illustration of the three grasp hypotheses images incorporated
    into the training set per hand. The blue triangles at the bottom
    denote positions of the two range sensors. (c-d) illustrate
    training images created using data from only one sensor.}
  \label{fig:trainingdata}
\end{figure}

\subsection{Creating the Training Set}

In order to create the training set, we obtain a set of objects that
have local geometries similar to what might be expected in the
field. In our work, we selected the set of 18 objects shown in
Figure~\ref{fig:trainingdata} (a). Each object was placed in front of
the robot in two configurations: one upright configuration and one on
its side. For each configuration (36 configurations total), let
$\mathcal{C}_1$ and $\mathcal{C}_2$ denote the voxelized point clouds
obtained from each of the two sensors, respectively, and let
$\mathcal{C}_{12} = \mathcal{C}_1 \cup \mathcal{C}_2$ denote the
registered two-view cloud.

The training data is generated as follows. First, we extract hand
hypotheses from the registered cloud, $\mathcal{C}_{12}$ using the
methods of Section~\ref{sect:finding_hypotheses}. Second, for each $h
\in H$, we determine whether it is a positive, negative, or
indeterminate by checking the conditions of
Definition~\ref{defn:nearantipodal}. Indeterminate hands are discarded
from training. Third, for each positive or negative hand, we extract
three feature descriptors: $HOG(I(\mathcal{C}_1,h))$,
$HOG(I(\mathcal{C}_2,h))$, and $HOG(I(\mathcal{C}_{12},h))$. Each
descriptor is given the same label and incorporated into the training
set. Over our 18 object training set, this procedure generated
approximately 6500 positive and negative labeled examples that were
used to train an SVM. We only did one round of training using this
single training set.

The fact that we extract three feature descriptors per hand in step
three above is important because it helps us to capture the appearance
of partial views in the training set. Figure~\ref{fig:trainingdata}
(b-d) illustrates the three descriptors for an antipodal hand. Even
though the closing region of this hand is relatively well observed in
$\mathcal{C}_{12}$, the fact that we incorporate
$HOG(I(\mathcal{C}_1,h))$ and $HOG(I(\mathcal{C}_2,h))$ into the
dataset means that we are emulating what {\em would} have been
observed if we only had a partial view. This makes our method much
more robust to partial point cloud information.

\subsection{Cross Validation}
\label{sect:crossvalidation}

We performed cross validation on a dataset derived from the 18
training objects shown in Figure~\ref{fig:trainingdata} (a). For each
object, we obtained a registered point cloud for two configurations
(total of 36 configurations). Following the procedure described in
this section, we obtained 6500 labeled features with 3405 positives 
and 3095 negatives. We did 10-fold cross validation on this
dataset using an SVM for the various Gaussian and polynomial kernels
available in Matlab. We obtained 97.8\% accuracy using a degree-three
polynomial kernel and used this kernel in the remainder of our
experiments. In the cross validation experiment
described above, the folds were random across the labeled pairs in the
dataset. This does not capture the effects of experiencing novel
objects or the expected performance when only single-view point clouds
are available. Therefore, we did the following. First, we trained the
system using the degree-three polynomial kernel on the 6500 labeled
examples as described above. Then, we obtained additional single-view
\begin{wrapfigure}{r}{0.4\textwidth}
\begin{center}
  \includegraphics[height=1.5in]{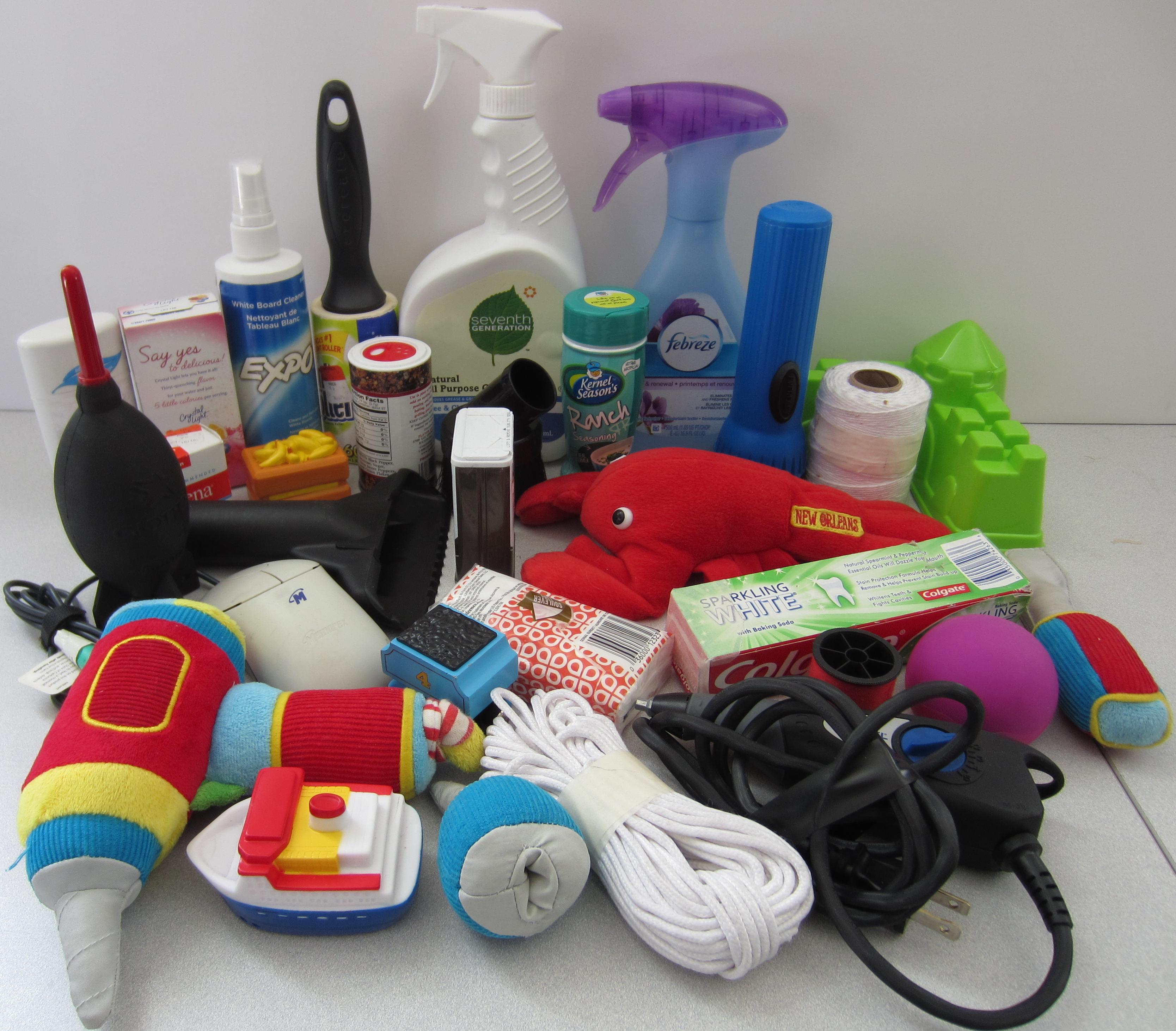}
\end{center}
  \caption{The 30 objects in our test set.}
  \label{fig:testset}
\end{wrapfigure}
point clouds for each of the 30 novel test objects shown in
Figure~\ref{fig:testset} (each object was presented in isolation) for
a total of 122 single-view points clouds. We used the methods
described in this section to obtain ground-truth for this
dataset. This gave us a total of 7250 labeled single-view hypotheses
on novel objects with 1130 positives and 6120 negatives. We obtained
94.3\% accuracy on this dataset. The fact that we do relatively well
in these cross validation experiments using a relatively simple
feature descriptor and without mining hard negatives suggests that our
approach to sampling hands and creating the grasp hypothesis image
makes the grasp classification task easier than it is in approaches
that do not use this kind of
structure~\cite{saxena_ijrr2008,jiang_icra2011,fischinger_iros2012,fischinger2013learning,lenz_rss2013}.

\section{Robot Experiments}
\label{sect:exps}

We evaluated the performance of our algorithms using the Baxter robot
from Rethink Robotics. We explore two experimental settings: when
objects are presented to the robot in isolation and when objects are
presented in a dense clutter scenario. We use the Baxter right arm
equipped with the stock two-finger Baxter gripper. A key constraint of
the Baxter gripper is the limited finger stroke: each finger has only
2 cm stroke. In these experiments, we adjust the finger positions such
that they are 3 cm apart when closed and 7 cm apart when open. This
means we cannot grasp anything smaller than 3 cm or larger than 7
cm. We chose each object in the training and test sets so that it
could be grasped under these constraints. Two-view registered point
clouds were created using Asus Xtion Pro range sensors (see
Figure~\ref{fig:stereo}). It should be possible for anyone with a
Baxter robot and the appropriate depth sensors to replicate any of
these experiments by running our ROS package at
\url{http://wiki.ros.org/agile_grasp}.

\subsection{Grasp Selection}
\label{sect:grasp_selection}

Since our algorithm typically finds tens or hundreds of potential
antipodal hands, depending upon the number of objects in the scene, it
is necessary to select one to execute. One method might be to select a
grasp on an object of interest. However, in this paper, we ignore
object identity and perform any feasible grasp. We choose a grasp to
attempt as follows. First, we sparsify the set of grasp choices by
clustering antipodal hands based on distance and orientation. Grasp
hypothesis that are nearby each other and that are roughly aligned in
orientation are grouped together. Each cluster must be composed of a
specified minimum number of constituent grasps. If a cluster is found,
then we create a new grasp hypothesis positioned at the mean of the
cluster and oriented with the ``average'' orientation of the
constituent grasps. The next step is to select a grasp based on how
easily it can be reached by the robot. First, we solve the inverse
kinematics (IK) for each of the potential grasps and discard those for
which no solution exists. The remaining grasps are ranked according to
three criteria: 1) distance from joint limits (a piecewise function
that is zero far from the arm joint limits and quadratic nearby the
limits); 2) distance from hand joint limits (zero far from the limits
and quadratic nearby limits); 3) workspace distance traveled by the
hand starting from a fixed pre-grasp arm configuration. These three
criteria are minimized in order of priority: first we select the set
of grasps that minimize Criterion \#1. Of those, we select those that
minimize Criterion \#2. Of those, we select the one that minimizes
Criterion \#3 as the grasp to be executed by the robot.

%Criterion \#1 is a piecewise quadratic measure that is zero far from
%the arm joint limits, but grows quadratically once the minimum
%distance to a arm joint limit falls below a specified threshold.
%This criterion is important because it is sometimes the case that
%reaching accuracy degrades in the vicinity of the arm joint limits
%(this is true for Baxter, for example).
%Criterion \#2 is distance from hand joint limits. This measure has the
%same mathematical form as the measure in Criterion \#1 except that it
%is evaluated for hand joints.
%This criterion is important because it penalizes grasps that are very
%close to the hand minimum or maximum aperture and are therefore less
%robust to perceptual errors.
%Criterion \#3 is the L2 distance in joint space.
%Minimizing this criterion biases grasp selection toward grasps that
%will be quickly executed.
%These three criteria are minimized in order of priority: first we
%select the set of grasps that minimize Criterion \#1. Of those, we
%select those that minimize Criterion \#2. Of those, we select those
%that minimize Criterion \#3.

%~\footnote{We calculate the ``average'' orientation using the
%  Eigenvectors of the covariance matrix of the constituent grasp
%  approach directions.}.

\subsection{Objects Presented in Isolation}

\begin{figure}[t]
\tiny
%\scriptsize 
\begin{center}
  \begin{tabular}{| l | c | c |@{\hskip 0.1in}| c | c | c | c | c |}
  \hline
%  \multirow{2}{*}{\textbf{Object}} & \textbf{number} & \multirow{2}{*}{{\bf A, 2V}} & \textbf{number} & \multicolumn{4}{|c|}{\textbf{Success Rate}} \\
  \multirow{2}{*}{\textbf{Object}} & \textbf{number} & {\bf Succ. Rate} & \textbf{number} & \multicolumn{4}{|c|}{\textbf{Success Rate}} \\
  \cline{5-8}
  & \textbf{of poses} & {\bf A, 2V} & \textbf{of poses} & \textbf{NC, 1V} & \textbf{NC, 2V} & \textbf{SVM, 1V} & \textbf{SVM, 2V} \\
  \hline
  Plush drill & 3 & 100.00\% & 6 & 50.00\% & 66.67\% & 100.00 & 66.67\% \\ \hline
  Black pepper & 3 & 100.00\% & 8 & 62.5\% & 62.50\% & 75.00 & 100.00\% \\ \hline
  Dremel engraver & 3 & 100.00\% & 6 & 33.33\% & 50.00\% & 66.67 & 100.00\% \\ \hline
  Sand castle & 3 & 100.00\% & 6 & 50.00\% & 33.33\% & 83.33 & 83.33\% \\ \hline
  Purple ball & 0 & NA & 6 & 66.67\% & 100.00\% & 83.33 & 100.00\% \\ \hline
  White yarn roll & 3 & 100.00\% & 8 & 87.50\% & 87.50\% & 87.50 & 75.00\% \\ \hline
  Odor protection & 0 & NA & 8 & 50.00\% & 87.50\% & 87.50 & 75.00\% \\ \hline
  Neutrogena box & 3 & 66.67\% & 8 & 25.00\% & 87.50\% & 87.50 & 87.50\% \\ \hline
  Plush screwdriver & 3 & 100.00\% & 6 & 83.33\% & 87.50\% & 83.33 & 100.00\% \\ \hline
  Toy banana box & 3 & 100.00\% & 8 & 100\% & 83.33\%  & 87.50 & 75.00\% \\ \hline
  Rocket & 3 & 100.00\% & 8 & 50.00\% & 87.50\% & 100.00 & 87.50\% \\ \hline
  Toy screw & 3 & 100.00\% & 6 & 100.00\% & 100.00\% & 83.33 & 100.00\% \\ \hline
  Lamp & 3 & 100.00\% & 8 & 62.50\% & 83.33\% & 87.50 & 87.50\% \\ \hline
  Toothpaste box & 3 & 66.67\% & 8 & 87.50\% & 100.00\% & 87.50 & 87.50\% \\ \hline
  White squirt bottle & 3 & 66.67\% & 8 & 25.00\% & 12.50\% & 75.00 & 87.50\% \\ \hline
  White rope & 3 & 100.00\% & 6 & 66.67\% & 83.33\% & 83.33 & 100.00\% \\ \hline
  Whiteboard cleaner & 3 & 100.00\% & 8 & 62.50\% & 75.00\% & 100.00 & 100.00\% \\ \hline
  Toy train & 0 & NA & 8 & 87.50\% & 100.00\% & 87.50 & 100.00\% \\ \hline
  Vacuum part & 3 & 100.00\% & 6 & 33.33\% & 66.67\% & 100.00 & 83.33\% \\ \hline
  Computer mouse & 0 & NA & 6 & 33.33\% & 33.33\% & 66.67 & 83.33\% \\ \hline
  Vacuum brush & 1 & 100\% & 6 & 50.00\% & 83.33\% & 66.67 & 50.00\% \\ \hline
  Lint roller & 3 & 100.00\% & 8 & 75.00\% & 75.00\% & 87.50 & 100.00\% \\ \hline
  Ranch seasoning & 3 & 100.00\% & 8 & 50.00\% & 75.00\% & 100.00 & 100.00\% \\ \hline
  Red pepper & 3 & 100.00\% & 8 & 75.00\% & 75.00\% & 100.00 & 100.00\% \\ \hline
  Crystal light & 3 & 100.00\% & 8 & 25.00\% & 37.50\% & 75.00 & 75.00\% \\ \hline
  Red thread & 3 & 100.00\% & 8 & 75.00\% & 100.00\% & 100.00 & 100.00\% \\ \hline
  Kleenex & 3 & 100.00\% & 6 & 33.33\% & 33.33\% & 83.33 & 83.33\% \\ \hline
  Lobster & 3 & 66.67\% & 6 & 16.67\% & 83.33\% & 66.67 & 83.33\% \\ \hline
  Boat & 3 & 100.00\% & 6 & 83.33\% & 100.00\% & 83.33 & 100.00\% \\ \hline
  Blue squirt bottle & 2 & 100\% & 8 & 25.00\% & 50.00\% & 75.00 & 62.50\% \\
  \hline
  \hline
  \textbf{Average} & & \textbf{94.67\%} & & \textbf{57.50\%} & \textbf{72.92\%} & \textbf{85.00\%} & \textbf{87.78\%} \\
  \hline
  \end{tabular}
 
  \caption{Single object experimental results. Algorithm variations
    are denoted as: A for antipodal grasps (see
    Section~\ref{sect:identify_antipodal}), NC for sampling without
    grasp classification (see Section~\ref{sect:finding_hypotheses}),
    and SVM for our full detection system.}
%     \centering
     \label{fig:per_obj_results_all}
\end{center}
\end{figure}

We performed a series of experiments to evaluate how well various
parts of our algorithm perform in the context of grasping each of the
30 test set objects (Figure~\ref{fig:testset}). Each object was
presented to the robot in isolation on a table in front of the
robot. We characterize three variations on our algorithm:

\begin{enumerate}
\item {\bf No Classification:} We assume that all hand hypotheses
  generated by the sampling algorithm (Algorithm~\ref{alg:gethypo})
  are antipodal and pass all hand samples directly to the grasp
  selection mechanism without classification as described in
  Section~\ref{sect:grasp_selection}.
\item {\bf Antipodal:} We classify hand hypotheses by evaluating the
  conditions of Definition~\ref{defn:nearantipodal} directly for each
  hand and pass the results to grasp selection.
\item {\bf SVM:} We classify hand hypotheses using the SVM and pass
  the results to grasp selection. The system was trained using the
  18-object training set as described in
  Section~\ref{sect:crossvalidation}.

\end{enumerate}

\noindent
In all scenarios, a grasp trial was considered a success only when the
robot successfully localized, grasped, lifted, and transported the
object to a box on the side of the table. We evaluate {\em No
  Classification} and {\em SVM} for single-view and two-view
registered points clouds over 214 grasps of the 30 test objects. Each
object was placed in between 6 and 8 systematically different
orientations relative to the robot.

Figure~\ref{fig:per_obj_results_all} shows the results. The results
for {\em No Classification} are shown in columns {\em NC, 1V} and {\em
  NC, 2V}. Column {\em NC, 1V} shows that with a point cloud created
using only one depth sensor, using the results of sampling with no
additional classification results in an average grasp success rate of
58\%. However, as shown in Column {\em NC, 2V}, it is possible to
raise this success rate to 73\% just by adding a second depth sensor
and using the resulting two-view registered cloud. The fact that we
obtain a grasp success rate as high as 73\% here is surprising
considering that the sample strategy employs rather simple geometric
constraints. This suggests that even simple geometric constraints can
improve grasp detection significantly. The results for {\em Antipodal}
are shown in the column labeled {\em A, 2V}. We did not evaluate this
variation for a one-view cloud because a two-view cloud is needed for
Definition~\ref{defn:nearantipodal} to find any near antipodal
hands. Compared to the other two approaches, {\em Antipodal} finds
relatively few positives. This is because this method needs to ``see''
two sides of a potential grasp surface in order to verify the presence
of a grasp. As a result, we were only able to evaluate this method
over three poses per object instead of six or eight. In fact, {\em
  Antipodal} failed to find any grasps at all for four of the 30
objects. Overall, {\em Antipodal} can be an effective way to detect
grasps (94.7\% grasp success rate), but since it is not robust to
occlusions at all, it is not very useful in practice. The results for
{\em SVM} are shown in columns {\em SVM, 1V} and {\em SVM, 2V}
(results for one-view and two-view point clouds,
respectively). Interestingly, there is not much advantage here to
adding a second depth camera: we achieve an 85.0\% success rate with a
one-view point cloud and an 87.8\% success rate with a two-view
registered cloud. Drilling down into these numbers, we find the
following three major causes of grasp failure: 1) approximately 5.6\%
of the grasp failure rate in both scenarios is due to collisions
between the gripper and the object caused by arm calibration errors or
collisions with observed or unobserved parts of the environment; 2)
approximately 3.5\% of the objects were dropped after a successful
initial grasp; 3) approximately 2.3\% of grasp failures in the
two-view case (3.7\% in the one view case) were caused by perceptual
errors. The striking thing about the causes of failure listed above is
that they are not all perceptual errors: if we want to improve beyond
the 87.8\% success rate, we need to improve performance in multiple
areas.

\begin{wrapfigure}{r}{0.32\textwidth}
  \begin{center}
    \includegraphics[height=1.15in]{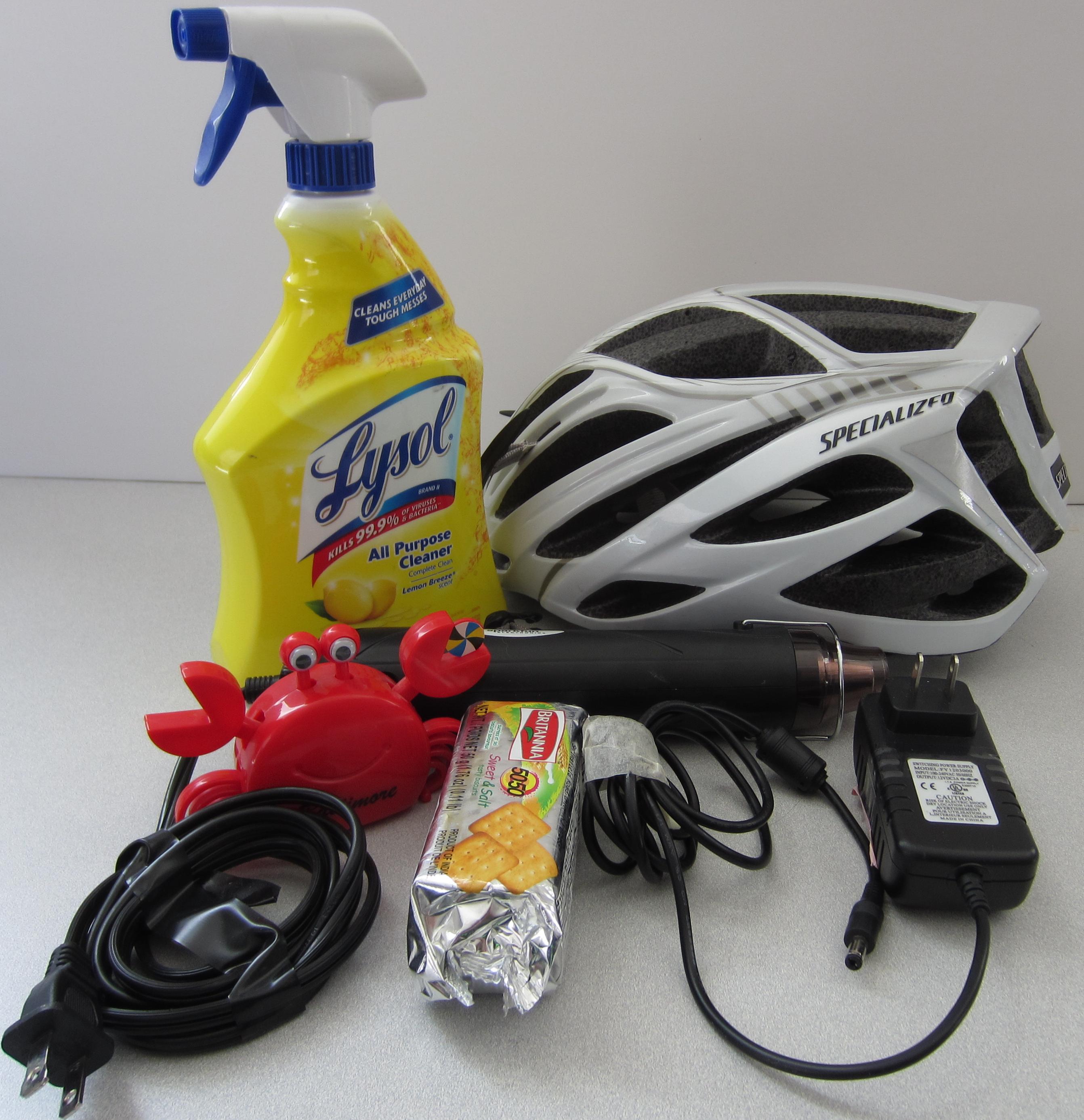}
%    \subfigure[]{\includegraphics[height=1in]{figs/hardtosee_results.png}}
  \end{center}
  \caption{Hard-to-see objects.}
  \label{fig:hardtosee}
\end{wrapfigure}

In the experiments described above, we eliminated seven objects from
the test set because they were hard to see with our depth sensor (Asus
Primesense) due to specularity, transparency, or color. We
characterized grasp performance for these objects separately by
grasping each of these objects in eight different poses (total of 56
grasps over all seven objects). Using \emph{SVM}, we obtain a 66.7\%
grasp success rate using a single-view point cloud and a 83.3\% grasp
success rate when a two-view cloud is used. This result suggests: 1)
our 87.8\% success rate drops to 83\% for hard-to-see objects; 2)
creating a more complete point cloud by adding additional sensors is
particularly important in non-ideal viewing conditions.

\subsection{Objects Presented in Dense Clutter}

%\begin{wrapfigure}{r}{0.4\textwidth}
%\end{wrapfigure}

\begin{figure*}
\begin{center}
  \subfigure[]{\includegraphics[height=1.6in]{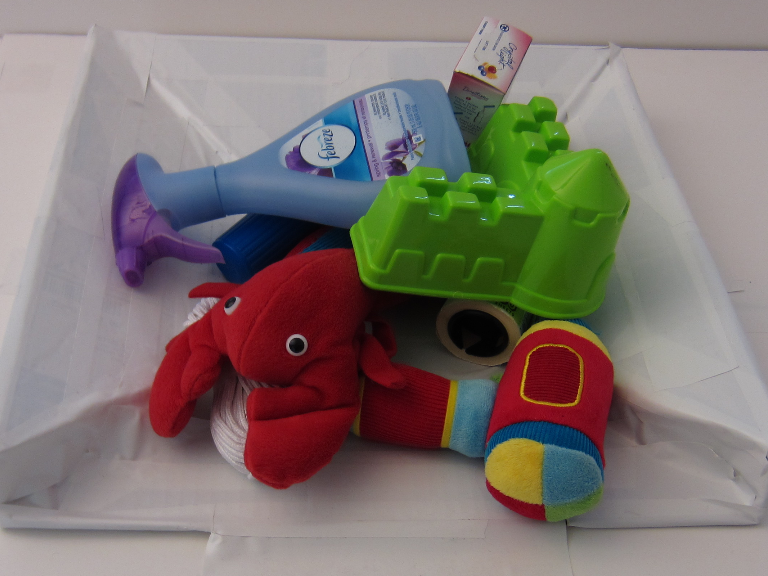}}
  \subfigure[]{\includegraphics[height=1.6in]{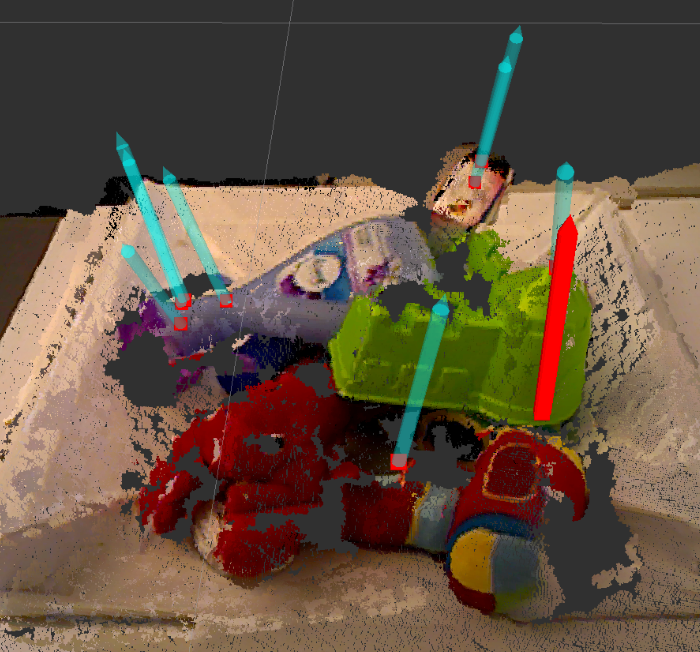}}
\end{center}
  \caption{Dense clutter scenario. (a) RGB image. (b) Output of our algorithm.}
  \label{fig:clutterexamples}
\end{figure*}

We also characterized our algorithm in dense clutter as illustrated in
Figure~\ref{fig:clutterexamples}. We created a test scenario where ten
objects are piled together in a shallow box. We used exactly the same
algorithm (i.e. \emph{SVM}) in this experiment as in the isolated object experiments. We
used a two-view registered point cloud in all cluttered scenarios. The
27 objects used in this experiment are a subset of the 30 objects used
in the single object experiments. We eliminated the computer mouse and
the engraver because they have cables attached to them that can get
stuck in the clutter. We also removed the vacuum brush because the
brush part cannot be grasped by the Baxter gripper in some
configurations due to the 3--7 cm aperture limits. At the beginning of
each run, we randomly selected 10 out of the 27 objects and placed
them in a small rectangular container. We then shook the container to
mix up the items and emptied it into the shallow box on top of the
table. We excluded all runs where the sandcastle landed upside down
because the Baxter gripper cannot grasp it in that configuration. A
run was terminated when three consecutive localization failures
occurred. In total, we performed 10 runs of this experiment.

%The purpose of the box is to prevent objects from being knocked off
%the table. The box has small ramps on the sides to minimize grasp
%errors caused by the robot attempting to grasp the box itself.

Over all 10 runs of this experiment, the robot performed 113
grasps. On average, it succeeded in removing 85\% of the objects from
each box. The remaining objects were not grasped because the system
failed to localize a grasp point three times in a row. Over all grasp
attempts, 73\% succeeded. The 27\% failure rate breaks down into the
following major failure modes: 3\% due to arm calibration errors; 9\%
due to perceptual errors; 4\% due to dropped objects following a
successful grasp; and 4\% due to collision with the environment. In
comparison with the isolation results, these results have a
significantly higher perceptual failure rate. We believe this is
mainly due to the extensive occlusions in the clutter scenario.

\section{Conclusion}

This paper proposes a new approach to localizing grasp points on novel
objects presented in clutter. Our main idea is to improve detection by
using geometric knowledge about good grasps. We first create a large
set of high quality grasp hypotheses by drawing samples that satisfy
simple, geometrically necessary conditions. We then use the geometry
of an antipodal grasp to create a large automatically labeled training
set that enables us to achieve high classification accuracy using an
SVM. If we omit the classification phase of this algorithm and
consider all samples to be good grasps, then we achieve an average
grasp success rate of 73\% when grasping objects presented in
isolation. This success rate is surprisingly high because the sampling
process only checks very simple necessary conditions on the presence
of a grasp. It suggests that our proposed geometry-based sampling
method is very effective. The average success rate increases to 87.8\%
when the sampled hypotheses are classified as antipodal grasps using
an SVM. When grasping novel objects presented in dense clutter, the
success rate drops to 73\% as a result of extensive occlusions. The
fact that performance drops so significantly in dense clutter suggests
that it is important to study the perceptual challenges unique to
dense clutter grasp scenarios. This paper is one of the first to
propose a systematic way of measuring grasp performance in dense
clutter. We hope to expand on this analysis of dense clutter in the
future. This system is available as a ROS package at
\url{http://wiki.ros.org/agile_grasp}.

%To our knowledge, we are the first to use grasping in very dense and
%randomly cluttered environments.

\bibliographystyle{plain}
{\tiny \bibliography{platt}}

\end{document}